\documentclass[12pt]{article}
\usepackage{balance}  % do not change this line -- unless you manually balance your last page

\newcommand{\und}[1]{\subsection*{#1}}
\newcommand{\uund}[1]{\paragraph{#1:}}
\newcommand{\droptext}[1]{{\em #1}}
\usepackage{amsmath}
\usepackage{marginnote}
\usepackage{hyperref}
\usepackage{amssymb}
\usepackage{mathtools}
\usepackage{wrapfig}
\usepackage{amsthm}
\usepackage[normalem]{ulem}
\usepackage{algorithm}
\usepackage{algpseudocode}
\usepackage{subcaption}
\usepackage{color}
\algtext*{EndWhile}% Remove "end while" text
\algtext*{EndIf}% Remove "end if" text
\algtext*{EndFor}% Remove "end for" text

\newcommand\fromrao[1]{}
\begin{document}

%\title{Synthesizing Explainable Behavior for Human-AI Collaboration}  % put your title here! 

\title{Challenges of Human-Aware AI Systems}

\author{Subbarao Kambhampati\thanks{This article is based on the AAAI 2018 Presidential Address that the author had the honor of delivering in New Orleans in February 2018. The video of the talk, along with the slides used, is available at \url{http://bit.ly/2tHyzAh}}\\
Arizona State University\\
rao@asu.edu}

\date{}
\maketitle

\begin{abstract}  % put your abstract here!
%Research in AI suffers from a longstanding ambivalence to humans-swinging as it does, between their replacement and augmentation. 
From its inception, AI has had a rather ambivalent relationship to humans---swinging between their augmentation and replacement. Now, as AI technologies enter our everyday lives at an ever increasing pace, there is a greater need for AI systems to work synergistically with humans. To do this effectively, AI systems must pay more attention to aspects of intelligence that helped humans work with each other---including 
%emotional and 
social intelligence. I will discuss the research challenges in designing such human-aware AI systems, including modeling the mental states of humans in the loop, recognizing their desires and intentions, providing proactive support, exhibiting explicable behavior, giving cogent explanations on demand, and engendering trust. I will survey the progress made so far on these challenges, and highlight some promising directions. I will also touch on the additional ethical quandaries that such systems pose. I will end by arguing that the quest for human-aware AI systems broadens the scope of AI enterprise, necessitates and facilitates true inter-disciplinary collaborations, and can go a long way towards increasing public acceptance of AI technologies.

% As AI technologies enter our everyday lives at an ever increasing pace, there is a greater need for AI systems to work synergistically with humans.  This requires AI systems to exhibit behavior that is explainable to humans. Synthesizing such behavior requires AI systems to reason not only with their own models of the task at hand, but also about the mental models of the human collaborators. Using several case-studies from our ongoing research, I will discuss how such multi-model planning forms the basis for explainable behavior.
\end{abstract}

%\fromrao{\sout{should we have a section called "what is human-aware AI and why is it needed"--sort of like what Barbara Grosz did?}}

%\fromrao{6000-8000 words; no equations. Current  version is \sout{2K} 3200 words. The AAAI transcript is 9K words}

%\fromrao2{after 1st: put the robot projection picture? Put a how to lie figure?}

%\fromrao2{after 1st: I should push anagha's work as explanation and explicability in the context of partial observability... what were you doing when I didn't see you?}

%\fromrao{**Should I put something about intention projection MSR demos?}

\fromrao{Better title for XAI community section; make some of those subsubsections; PUT fetch robot showing intention prjection?}

%\fromrao{FIgures: \sout{Remove the "fake news" part of the old woman; add the micro-worlds up front; swap the robot walking corridors figure;} change the HuMM figure}

\fromrao{acks: thank Tathagata for figures}

%\fromrao{Basic idea: \sout{Keep the AAMAS stuff as our work--extending some descriptions; put the front motivation about why human-aware AI was ignored for long.} May be use the NYT op-ed. }

%\fromrao{pictures to put: \sout{the agent architecture picture; the multiple models picture}; \sout{pictures of Robot going in the hallway?} }

\fromrao{Add more outside cites}

\fromrao{check unicode -- etc}

%use rao robot pic in bio

\section*{Introduction}
Artificial Intelligence, the discipline we all call our intellectual home, is suddenly having a rather huge cultural moment. It is hard to turn anywhere without running into mentions of AI technology and hype about its expected positive and negative societal impacts. AI has been compared  to fire {\em and} electricity, and commercial interest in the AI technologies has sky rocketed. 
%Companies and investors are pouring money into the field. 
Universities -- even high schools -- are rushing to start new degree programs or colleges dedicated to AI. Civil society organizations are scrambling to understand the impact of AI technology on humanity, and governments are competing to encourage or regulate AI research and deployment. 

There is considerable hand-wringing by pundits of all stripes on whether in the future, AI agents will get along with us or turn on us. Much is being written about the need to make AI technologies safe and delay the ``doomsday." I believe that as AI researchers, we are not (and cannot be) passive observers.  It is {\em our} responsibility to design agents that can and will get along with us. Making such {\em human-aware} AI agents, however poses several foundational research challenges that go beyond simply adding user interfaces \textit{post facto}. I will argue that addressing these challenges  
%also 
broadens the scope of AI in fundamental ways. 

%\marginnote{Making such {\em human-aware} AI agents, however poses several foundational research challenges that go beyond simply adding user interfaces \textit{post facto}. I will argue that addressing these challenges  also broadens the scope of AI in fundamental ways. }[3cm]

%\begin{wrapfigure}{r}{3in}
\begin{figure}
\begin{center}
\includegraphics[width=5in]{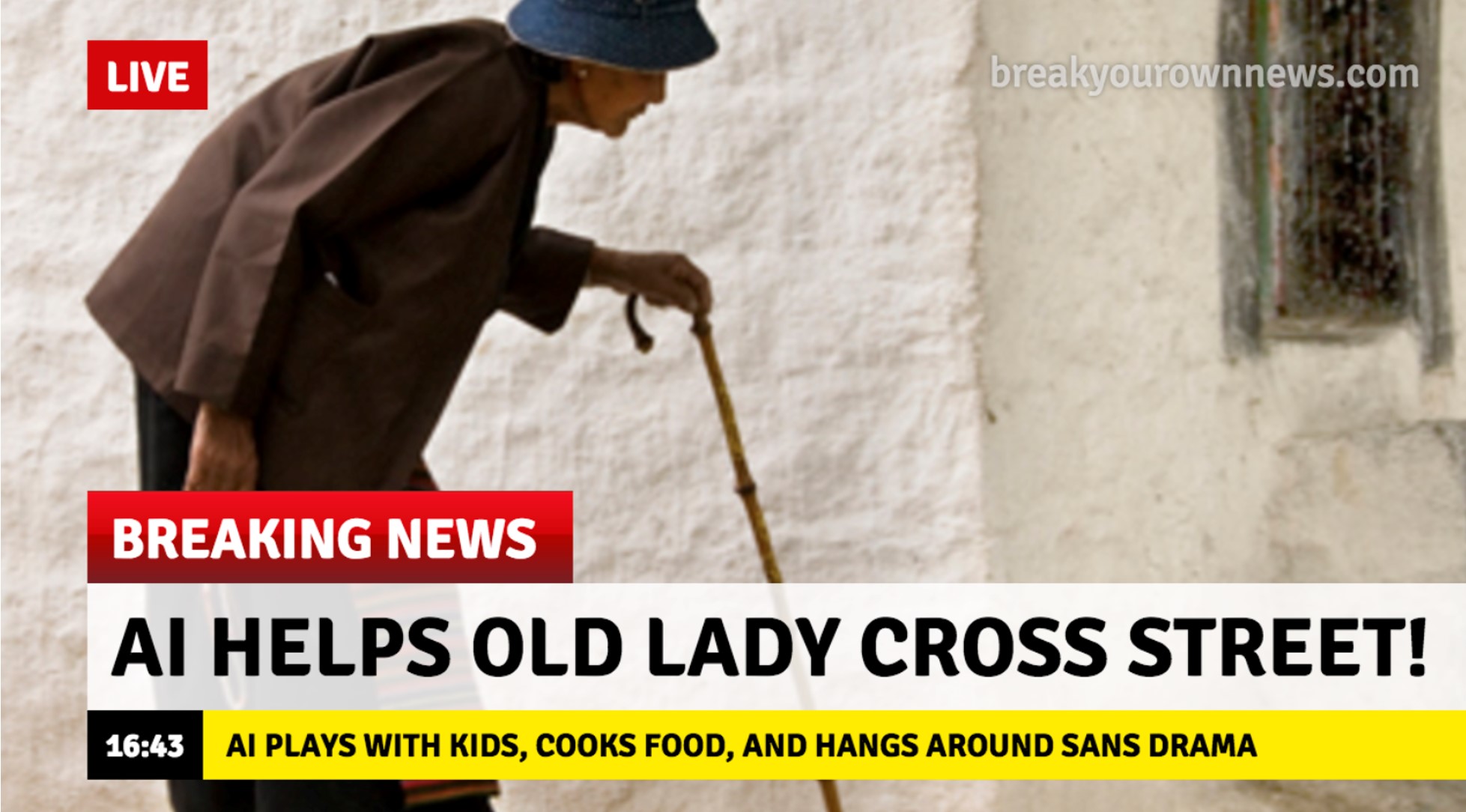}
%{lady-cross-street.jpg}
\end{center}
\caption{\em We should build a future where AI systems can be our quotidian partners}
%\end{wrapfigure}
\end{figure}

\subsection*{The need for Human-Aware AI Systems}
My primary aim in this talk is to call for an increased focus on human-aware AI systems---goal directed autonomous systems that are capable of effectively interacting, collaborating and teaming with humans.\footnote{In a way, it thus  follows in the footsteps of Barbara Grosz's AAAI Presidential Address \cite{barbara-presidential}, which talked about collaborative systems.} Although developing such systems seems like a rather self-evidently fruitful enterprise, and popular imaginations of AI, dating back to HAL,  almost always assume we already do have human-aware AI systems technology, little of the actual energies of the AI  research community have gone in this direction. 
%It is worth asking why

From its inception, AI has had a rather ambivalent relationship to humans---swinging between their augmentation and replacement. Most high profile achievements of AI have either been far away from the humans---think Spirit and Opportunity exploring Mars; or in a decidedly adversarial stance with humans, be it Deep Blue, AlphaGo or Libatus. Research into effective ways of making AI systems {\em interact, team and collaborate with  humans}  has received significantly less attention. It is perhaps no wonder that many lay people have fears about AI technology! 

This state of affairs is a bit puzzling given the rich history of early connections between AI and psychology. 
Part of the initial reluctance to  work on these issues  had to do with the worry that focusing on AI systems working with human might somehow dilute the grand goals of the AI enterprise, and might even lead to  temptations of  ``cheating,'' with most of the intelligent work being done by the humans in the loop. After all, prestidigitation has been a concern  since the original mechanical turk.  Indeed, much of the early work on human-in-the-loop AI systems mostly focused on using humans as a crutch for making up the limitations of the AI systems \cite{allen1994mixed}. In other words, early AI had humans be ``AI-aware" (rather than AI be ``human-aware").

Now, as AI systems are maturing with increasing capabilities, the concerns about them depending on humans as crutches are less severe. I would also argue that  focus on humans in the loop doesn't dilute the goals of AI enterprise, but in fact broadens them in multiple ways. After all, evolutionary theories tell us that humans may have developed the brains they have,  not so much to run away from the lions of the savanna or tigers of Bengal but rather to effectively cooperate and compete with each other. Psychological tests such as the Sally Anne Test
%\footnote{https://en.wikipedia.org/wiki/Sally%E2%80%93Anne_test} 
\cite{sally-anne-test} demonstrate the importance of such social cognitive abilities in the development of collaboration abilities in children. 
\fromrao{is this too strong?}

%\fromrao{Barbara Grosz talked about collaborative systems. Eric Horvitz. Cynthia}

Some branches of AI, aimed at specific human-centric applications, such as intelligent tutoring systems\cite{kurt-its1}, and social robotics \cite{cynthia-book,cynthia-ros,scaz-tom}, did focus on the challenges of human-aware AI systems for a long time. It is crucial to note however that human-aware AI systems are needed in a much larger class of quotidian applications beyond those. These include human-aware AI assistants for many applications where humans continue to be at the steering wheel, but will need naturalistic assistance from AI systems---akin to what they can expect from a smart human secretary.
%Should there be a mention of future of work? 
\droptext{Increasingly, as AI systems become common-place, human-AI interaction will be the dominant form of human-computer interaction} \cite{weld-chi}.

%It is for this reason that 
For all these reasons and more, human-aware AI has started coming to the forefront of AI research of late.  Recent road maps for AI research, including the 2016 JASON report\footnote{https://fas.org/irp/agency/dod/jason/ai-dod.pdf} and the 2016 White House OSTP report\footnote{https://obamawhitehouse.archives.gov/sites/default/files/whitehouse\_files/\\
microsites/ostp/NSTC/national\_ai\_rd\_strategic\_plan.pdf} emphasize the need for research in human-aware AI systems. The 2019 White House list of strategic R\&D priorities for AI lists ``developing effective methods for human-AI collaboration" at the top of the list of priorities\footnote{https://www.whitehouse.gov/wp-content/uploads/2019/06/National-AI-Research-and-Development-Strategic-Plan-2019-Update-June-2019.pdf}. Human-Aware AI was the special theme for the 2016 International Joint Conference on AI (with the tagline ``{\em why intentionally design a dystopian future and spend time being paranoid about it?}); it has been a  special track at AAAI since 2018. 
%When I was the program chair of IJCAI 2016 (International Joint Conference on Artificial Intelligence), we chose human-aware AI to be the special theme of the conference (with the tagline ``{\em why intentionally design a dystopian future and spend time being paranoid about it?}).
%One of the thematic pillars of Partnership for AI is collaborations between people and AI systems. 

%a concern that people can levy is that the important part from HCI view is computer systems that are "learning" based instead of programming based

%task vs. team level intervention

%IJCAI theme etc. 

%the COBOT project does bring a way of complementing.. 

\begin{figure*}
\centering
\includegraphics[width=\textwidth]{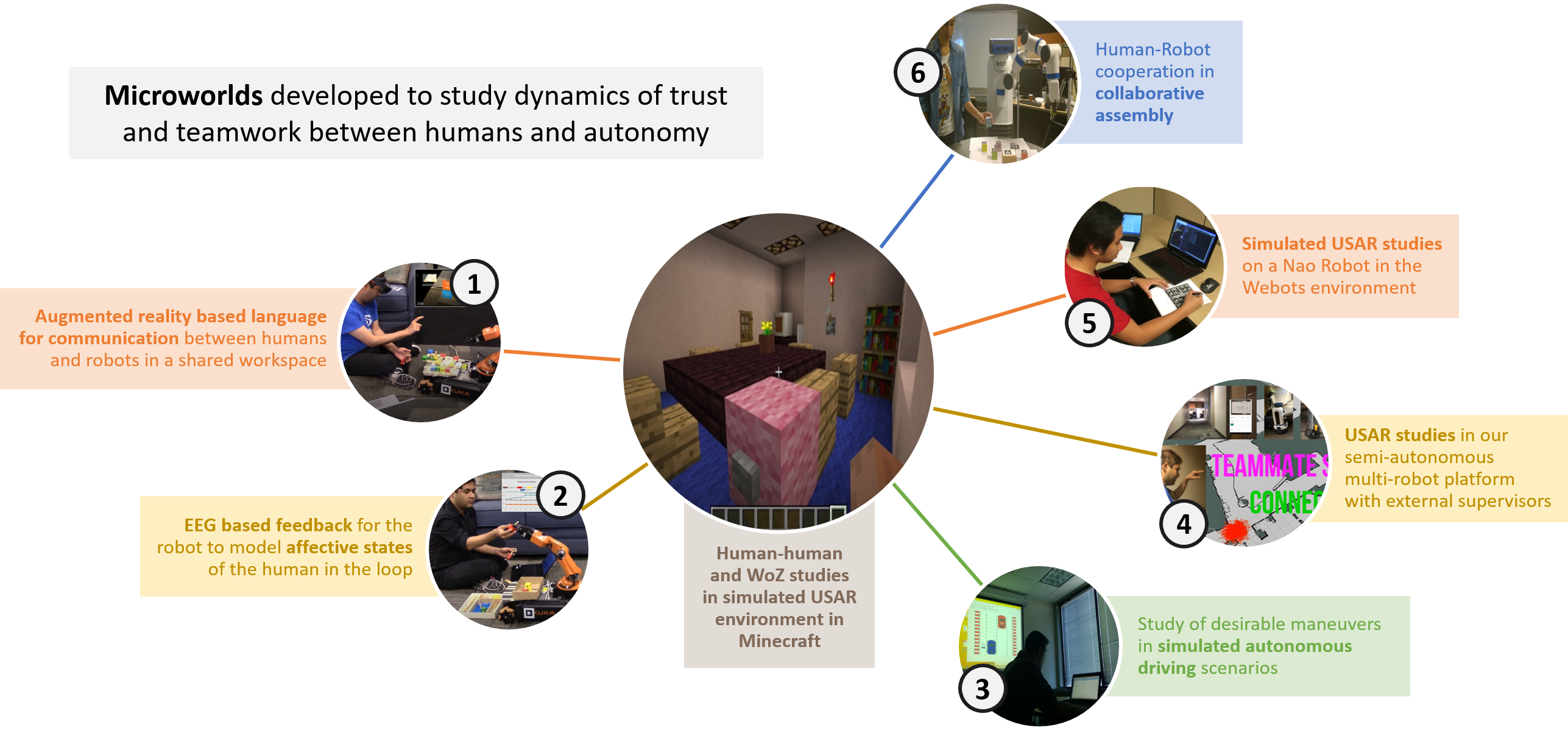}
\caption{\em Test beds developed to study the dynamics of trust and teamwork between autonomous agents and their human teammates.}
\label{fig:testbeds}
\end{figure*}

%naturalistic decision-making cites
\subsection*{How do we make AI agents Human-Aware?}

\begin{figure}[t]
\centering
\includegraphics[width=5in]{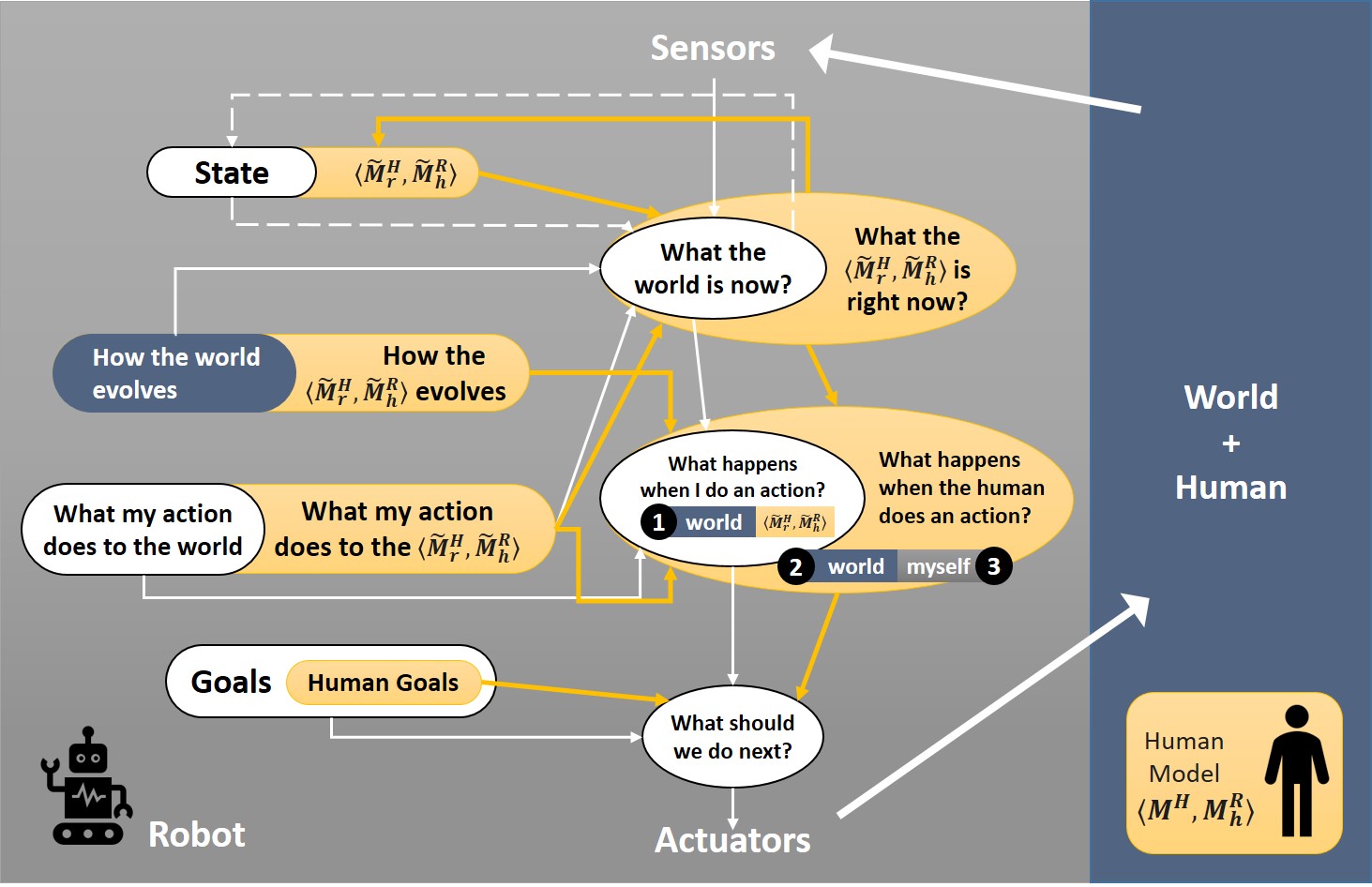}
%{newagent}
\caption{\em Architecture of an intelligent agent that takes human mental models into account. All portions in yellow are additions to the standard agent architecture, that are a result of the agent being human-aware. $M^R_h$ is the mental model the human has of the AI agent's goals and capabilities and $M^H_r$ is the (mental) model the  AI agent has of the human's goal and capabilities (see the section on Mental Models in Human-Aware AI)}
\label{agent}
\label{fig:newagent}
\end{figure}

When two humans collaborate to solve a task, both of them will develop approximate models of the goals and capabilities of each other (the so called ``theory of mind''), and use them to support fluid team performance. AI agents interacting with humans -- be they embodied or virtual -- will also need to take this implicit mental modeling into account. This certainly poses several research challenges. Indeed, it can be argued that acquiring and reasoning with such models changes almost every aspect of the architecture of an intelligent agent. As an illustration, consider the architecture of an intelligent agent that takes human mental models into account shown in Figure~\ref{fig:newagent}. Clearly most parts of the agent architecture -- including state estimation, estimation of the evolution of the world, projection of its own actions, as well as the task of using all this knowledge to decide what course of action the agent should take -- are all critically impacted by the need to take human mental models into account. This in turn gives rise to many fundamental research challenges. In \cite{cognitive-robot-teaming} we attempt to provide a survey of these challenges. 
Rather than list the challenges again here, in the rest of this article, I will use the ongoing work in our lab to illustrate some of these challenges as well as our current attempts to address them.\footnote{A longer bibliography of work related to human-aware AI from other research groups can be found at \url{http://rakaposhi.eas.asu.edu/cse591} as part of a graduate seminar at ASU on the topic.}  Our work has focused on the challenges of human-aware AI in the context of human-robot interaction scenarios \cite{chakraborti2018projection}, as well as human decision support scenarios \cite{sengupta2017radar}. Figure~\ref{fig:testbeds} shows some of the test beds and micro-worlds we have used in our ongoing work.

%present a framework of solutions to some of these challenges. 

%\fromrao{Put the paragraph headings into subsections}
%\fromrao{}

\und{Mental Models in Human-Aware AI}

In our ongoing research, we address the following central question in designing human-aware AI systems: ``{\em What does it take for an AI agent to show explainable behavior in the presence of humans?}. Broadly put, our answer is this: {\em :
To synthesize explainable behavior, AI
agents need to go beyond planning with their own
models of the world, and take into account the
mental model of the human in the loop. The
mental model here is not just the goals and
capabilities of the human in the loop, but
includes the human’s model of the AI agent’s
goals/capabilities.}

%In order for the AI agents to show behavior that makes sense to the human, they need to  go beyond planning with their own models of the world, and take into account the mental model of the human in the loop. The mental model here is not just the goals and capabilities of the humans in the loop, but includes the human's model of the AI agent's goals/capabilities. 
%show the new intelligent agent picture saying how everything changes?

\begin{figure}
\begin{center}
\begin{tabular}{c|c}
\includegraphics[width=2.5in]{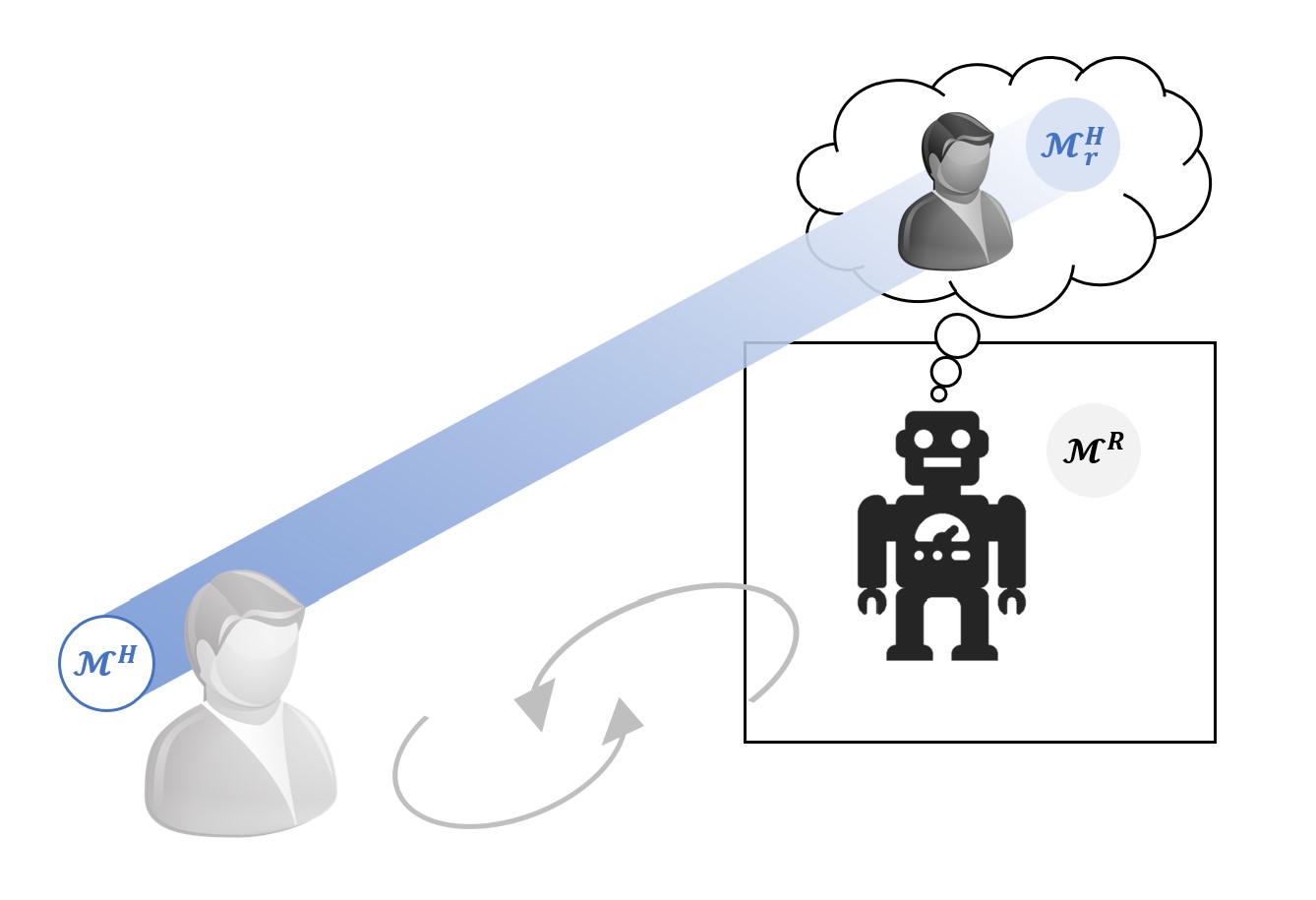} &
\includegraphics[width=2.5in]{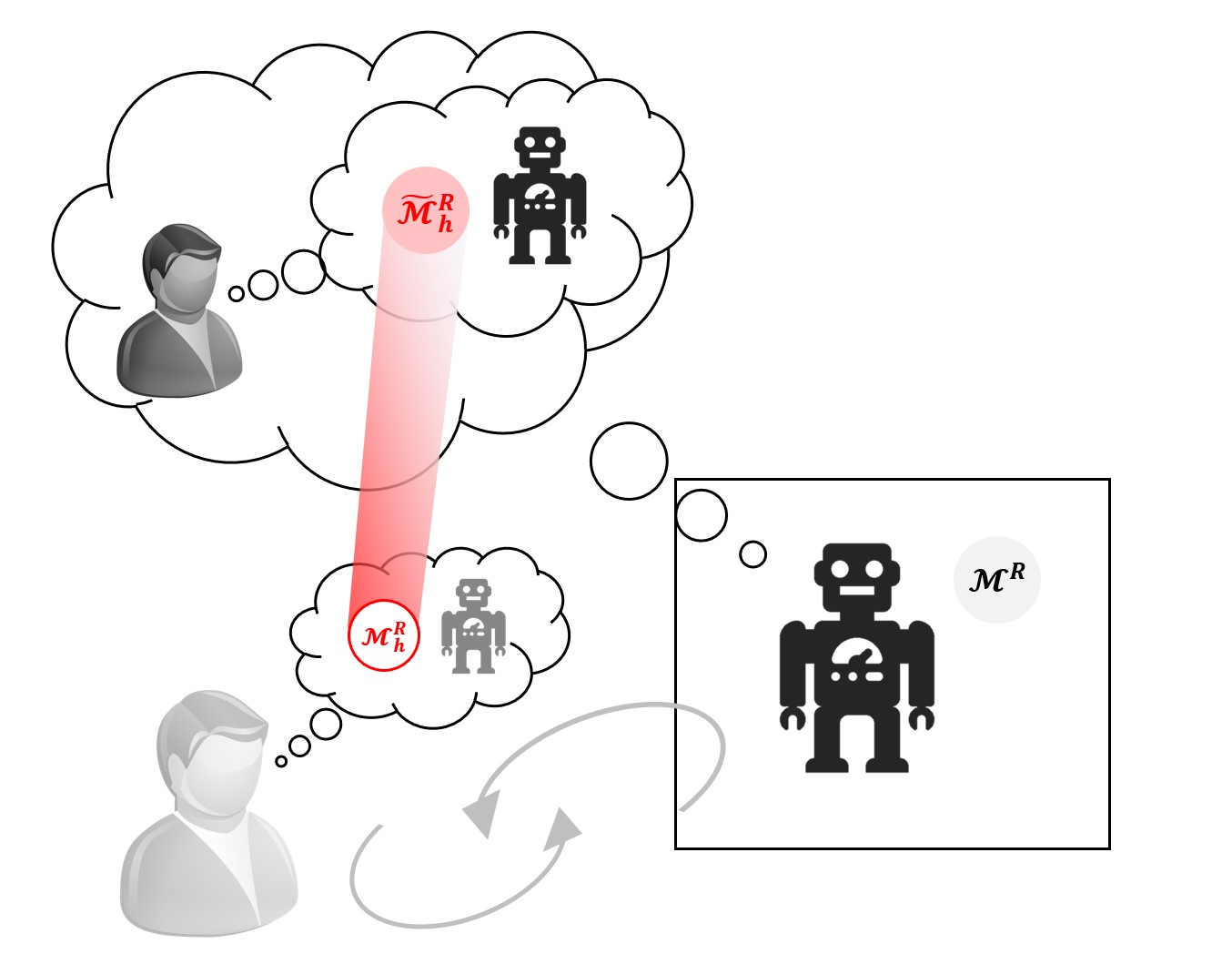}
\end{tabular}
\end{center}
\caption{\em Use of different mental models in synthesizing explainable behavior. (Left) The AI system can use its estimation of human's mental model, $M^H_r$, to take into account the goals and capabilities of the human thus providing appropriate help to them. (Right) The AI system can use its estimation of human's mental model of its capabilities $M^R_h$ to exhibit explicable behavior and to provide explanations when needed.}
%\end{wrapfigure}
\end{figure}

%In particular, it is useful to distinguish between these models: 
Let $M^R$ and $M^H$ correspond to the actual goal/capability models of the AI agent and human. To support collaboration, the AI agent needs an approximation of $M^H$, we will call it $\widetilde{M}^H_r$, to take into account the goals and capabilities of the human. The AI agent also needs to recognize that the human will have a model of its goals/capabilities $M^R_h$, and needs an approximation of this, denoted $\widetilde{M}^R_h$.  All phases of the ``sense--plan--act" cycle of an intelligent agent will have to change appropriately to track the impact on these models (as shown in Figure~\ref{fig:newagent}. Of particular interest to us in this article is the fact that synthesizing explainable behavior  becomes a challenge of supporting planning in the context of these multiple models. 

%to keep track of impact of the AI agent's decisions on the 
\fromrao{should we put the XAI relation here??}

In the following, we will look at some specific issues and capabilities provided by such human-aware AI agents. 
A note on the model representation: In much of our work, we have used relational precondition-effect models. We believe however that our frameworks can be readily adapted to other model representations; e.g. \cite{modelfree}.

%There is the learning angle.. 
%should mention the AAMAS/IROS papers on serendipity and resource contention

%\fromrao{some version of the AAMAS picture will be needed to go though this part}

\uund{Proactive help}
Left to itself, the AI agent will use $M^R$ to synthesize its behavior. When the agent has access to $\widetilde{M}^H_r$, we show how it can use that model to plan behaviors that proactively help the human user--either by helping them complete their goals (c.f. \cite{serendipity}) or avoiding resource contention with them (c.f. \cite{chakraborti2016planning}). 
%Why isn't there a widetilde in the robot's view of human

%Autonomous agents don't need explicability...or explanation.. 

\uund{Explicability}
When the agent has access to $\widetilde{M}^R_h$, it can use that model to ensure that its behavior is explainable.  We start by looking at generation of \textit{explicable behavior}, which requires the AI agent to not only consider the constraints of its model $M^R$, but also ensure that its behavior is in line with what is expected by the human. We can formalize this as finding a plan $\pi$ that trades off the optimality with respect to $M^R$ and ``distance'' from the plan $\pi'$ that would be expected according to $\widetilde{M}^R_h$. This optmization can be done either in a model-based fashion, where the distances between $\pi$ and $\pi'$ are explicitly estimated (c.f. \cite{Kulkarni:2019:EMD}) or in a model-free fashion, where the distance is indirectly estimated with the help of a learned ``labeling'' function that evaluates how far $\pi$ is from the expected plan/behavior (c.f. \cite{exp-yz}). Our notion of explicability here has interesting relations to other notions of interpretable robot behavior considered in AI and robotics communities; we provide a critical comparison of this landscape in \cite{chakraborti2019landscape}.

%I could say little about the methods used model-free and model-based

%inexplicability can occur also because of the wrong approximation.. but then explanation won't work either.. 
%****We need to consider the possibility that M^R_h contains some additional preferences human has on the robot's behavior. [This is still different from what they themselves want to do.. ]
%%What robot wants to change is M^R_h, but it reasons about the explanation with its own approximation \widetilde{M}^R_h
%Instead of constraining its behavior to be explicable to the human

\begin{figure}
\begin{center}
\includegraphics[width=6in]{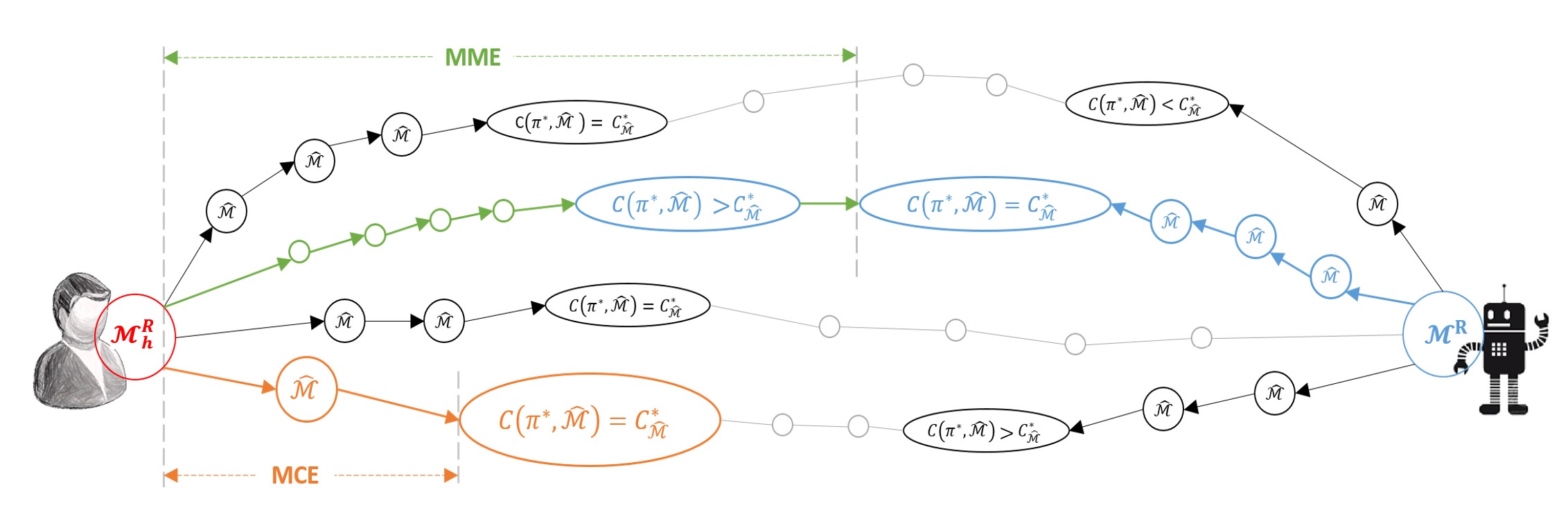}
%{lady-cross-street.jpg}
\end{center}
\caption{\em Computing explanations as model reconciliation involves a search in the space of the models. Here the AI agent's model $M^R$ is on the right end, and the human's model of the AI agent's capabilities, $M^R_h$ is on the left. The search transitions corresponds to model changes (for planning models, these might be addition/deletion of preconditions and effects). As discussed in \cite{explain}, the explanation process involves the AI agent searching for the minimal set of changes to reconcile the human's model to the actual model of the AI agent {\em in the context of the current problem}) for details)}
\label{fig:model-search}
%\end{wrapfigure}
\end{figure}

\uund{Explanation}
In some cases, $\widetilde{M}^R_h$ might be so different from $M^R$ that it will be too costly or infeasible for the AI agent to conform to those expectations. In such cases, the agent needs to provide an {\em explanation} to the human (with the aim of making its  behavior more explicable). We view explanation as a process of ``{\em model reconciliation,}'' specifically the process of helping the human bring $M^R_h$ closer to $M^R$. While a trivial way to accomplish this is to send the whole of $M^R$ as the explanation, in most realistic tasks, this will be both costly for the AI agent to communicate, and more importantly, for the human agent to comprehend. Instead, the explanation should focus on minimal changes $\mathcal{E}$ to $M^R_h$, such that the robot behavior $\pi$ is explicable with respect to $M^R_h + \mathcal{E}$, thus in essence making the behavior interpretable to human in light of the explanation. In \cite{explain} we show that computing such explanations can be cast as a \textit{meta search} in the space of models spanning $M^R$ and $\widetilde{M}^R_h$ (which is the AI agent's approximation of $M^R_h$); see Figure~\ref{fig:model-search}. We also provide methods to make this search more efficient, and discuss a spectrum of explanations with differing properties that can all be computed in this framework.  

\begin{figure}
\centering
\includegraphics[width=4in]{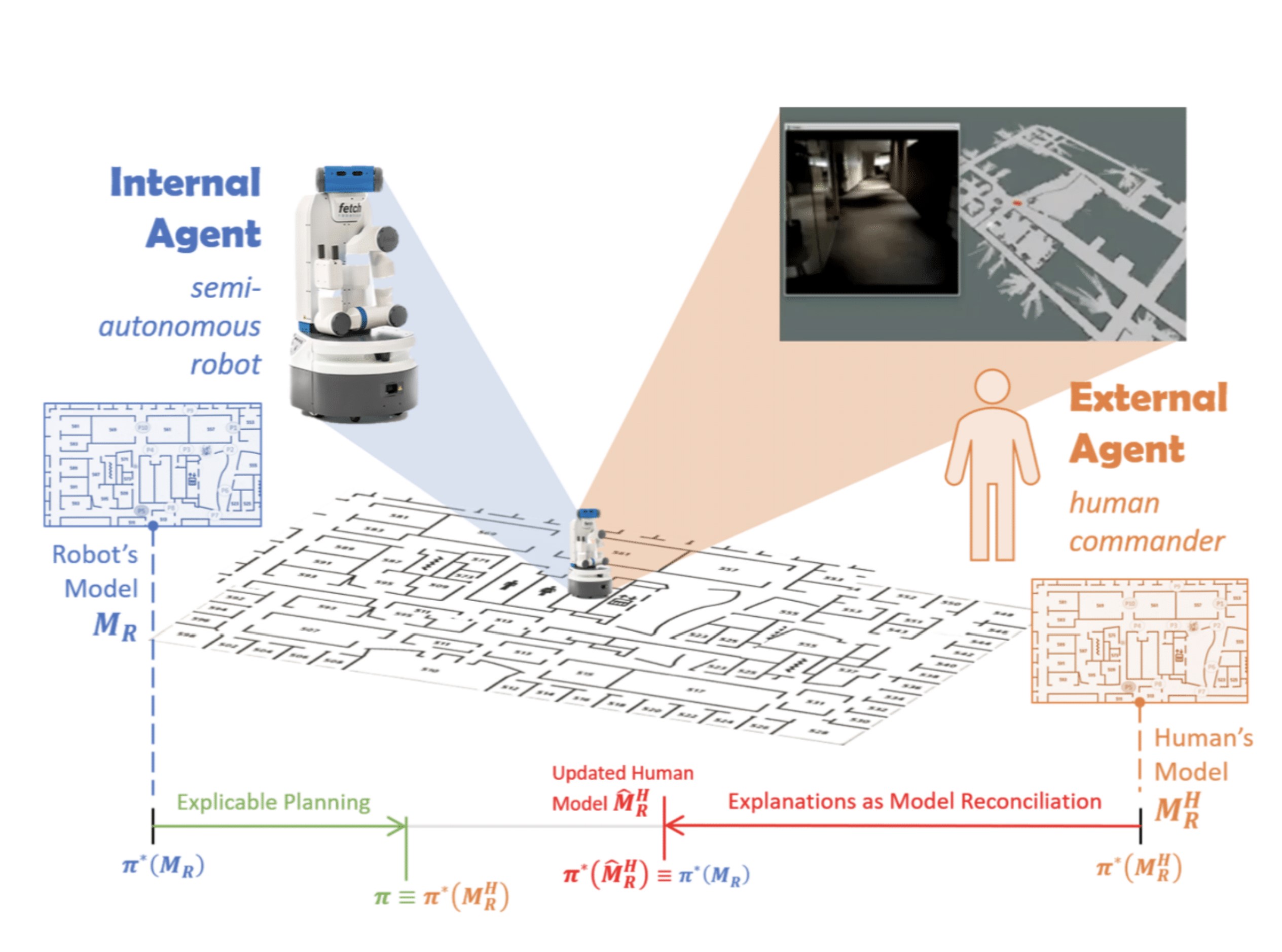}
\caption{\em A simplified urban search and rescue scenario where human and AI agents collaborate.}
\label{usr-setup}
\end{figure}

\begin{figure}
\begin{center}
\begin{tabular}{c|c}
\includegraphics[width=2.5in]{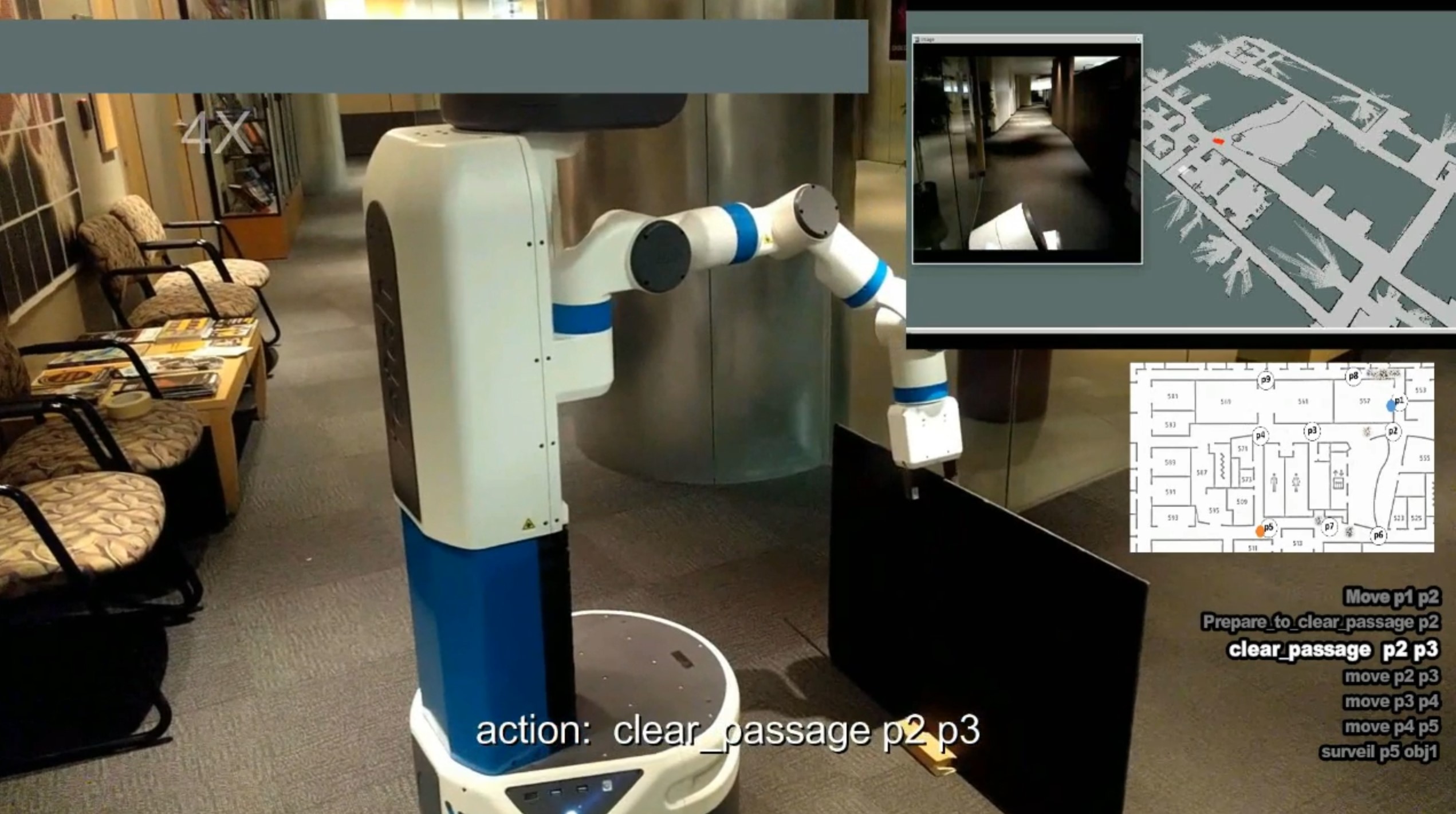} &
\includegraphics[width=2.5in]{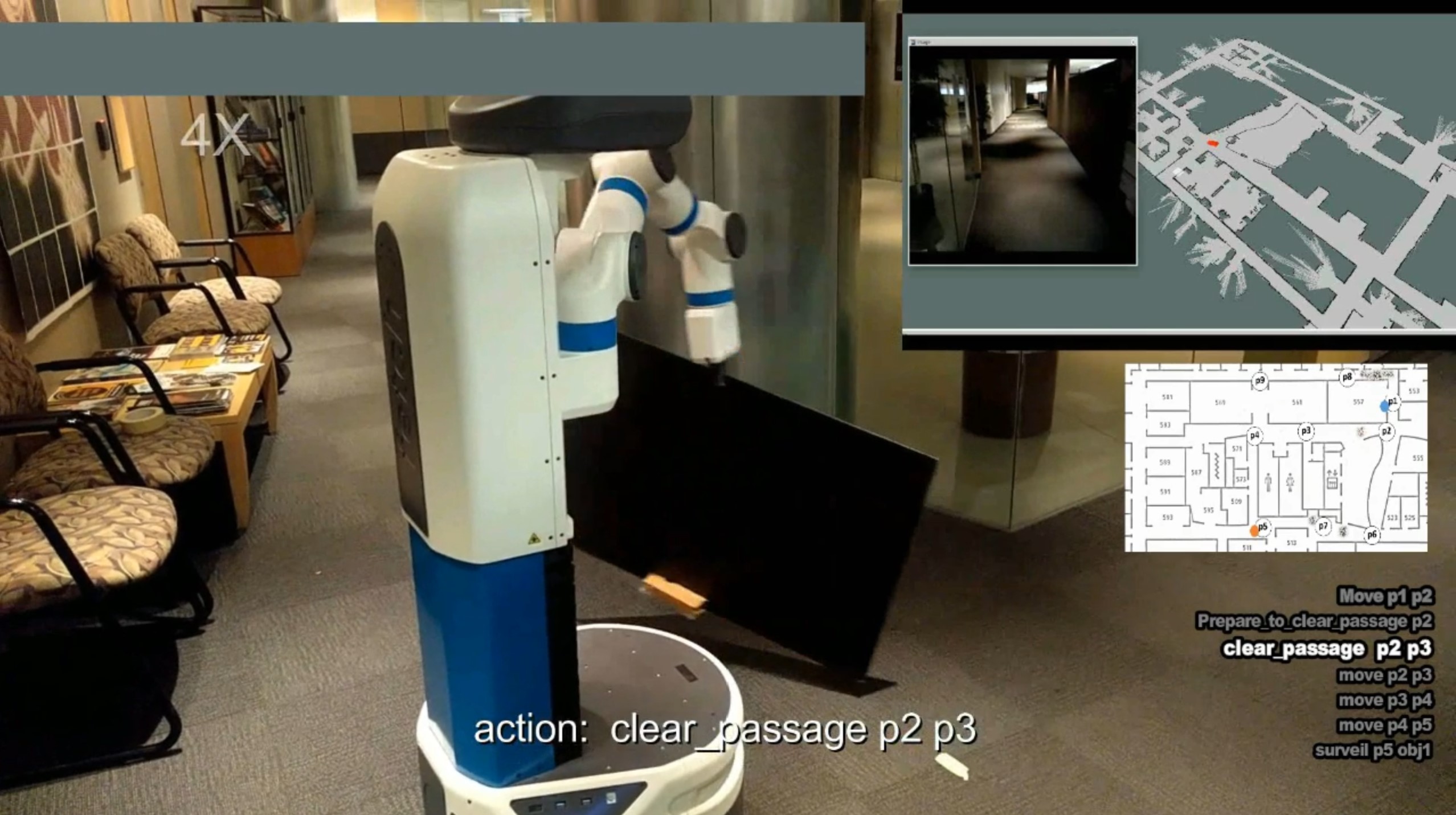}
\end{tabular}
\end{center}
\caption{\em In the case of explicable behavior, the AI agent behaves in the way the human commander expects it to believe (based on the commander's model $M^R_h$. This can be costly (and sometimes even infeasible) for the AI agent---as it is here, for example, where the robot has to {\em remove} the obstacle and clear the path so it can navigate it.}
\label{usr-explicability}
%\end{wrapfigure}
\end{figure}

\fromrao{Should the example go first?}
\uund{Example}  To illustrate the ideas of explicability and explanation in a concrete scenario, consider a simplified ``urban search and rescue" scenario depicted in Figure~\ref{usr-setup}. Here the human is in a commander's role, and is not at the scene of the search and rescue.  The robot (AI agent) -- which is at the scene -- collaborates with the human to search for the injured.  Both agents start with the same map of the environment. However, as the robot explores the environment, it might find that some of the pathways are blocked because of fallen debris.  In the example here, the robot realizes that the shortest path -- as expected by the human -- is blocked (see the black ``obstacle" on the left in Figure~\ref{usr-explicability}).  At this point, the robot has two choices. It can be explicable---by going through the path that the human expects. This will however involve the robot clearing the path by removing the obstacle (see Figure~\ref{usr-explicability} right side). Alternately, it can take the path that is optimal to it given the new map. In this case, the robot's explanation (to the possibly perplexed) human commander involves communicating the salient differences between $M^R_h$ and $M^R$ (see the message on top left in  Figure~\ref{usr-explicability}).

%\fromrao{Should I put the model-space search tree thing too?--decide while reading}

\begin{figure}
\centering
\includegraphics[width=4in]{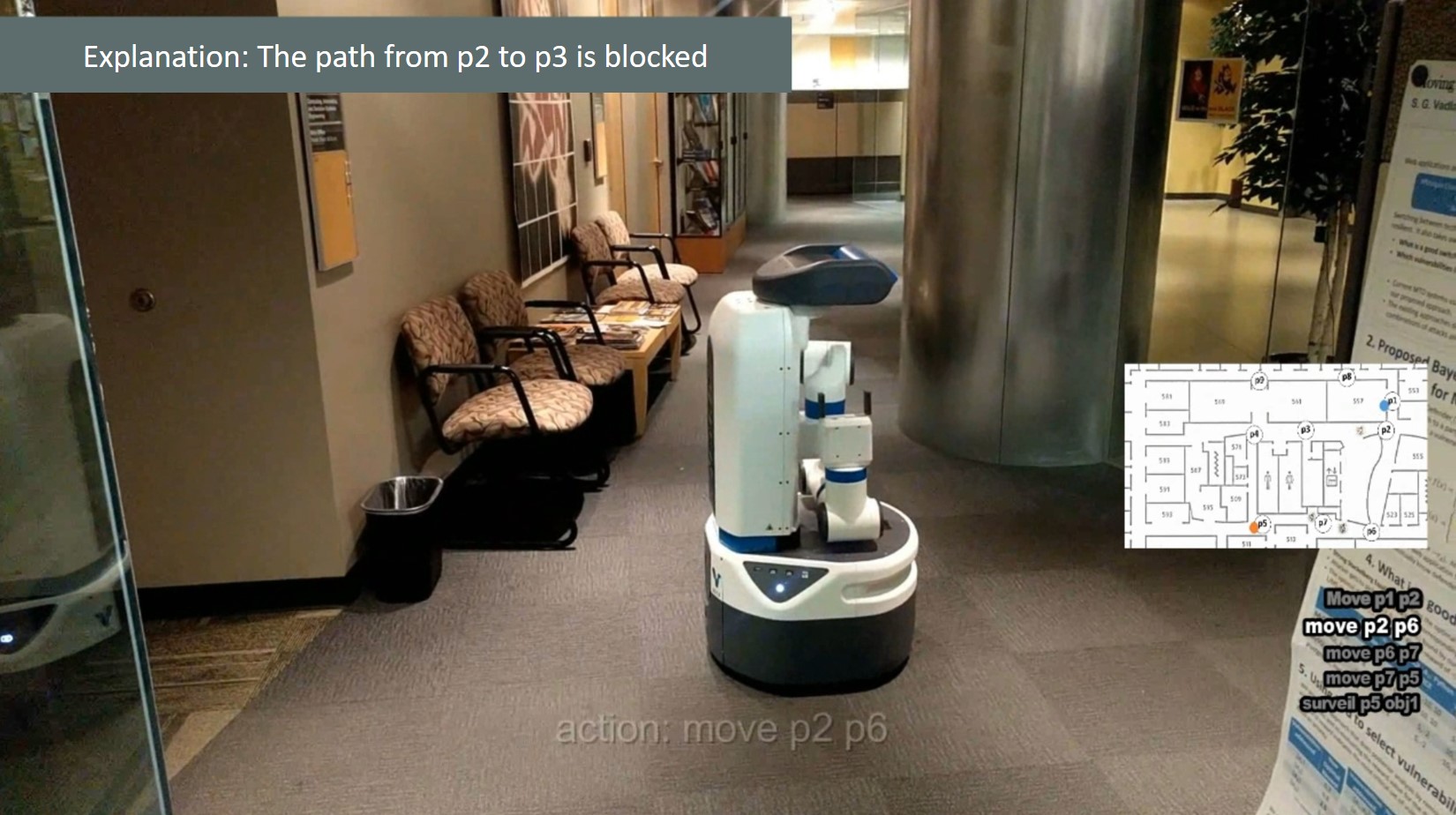}
%{usr-explaining.jpg}
\caption{\em When explicable behavior is too costly or infeasible, the AI agent can take the path that is optimal to it (given that the original shortest path is blocked), and provide an explanation. The explanation involves communicating the model differences between $M^R_h$ and $M^R$. For our case, this is just communicating that the shortest path is blocked (see the message at the top left)}
\label{usr-explanation}
\end{figure}

%\fromrao{SAY Neither is explanation a soliloquy... systems can't just dump their internal processing... e.g. certificates}

%Tim Miller citation

%use explainable instead of interpretable? 
\uund{Balancing Explicability \& Explanation}
While the foregoing presented showing explicable behavior and giving explanation as two different ways of exhibiting explainable behavior, it is possible to balance the trade-offs between them. In particular, given a scenario where $\pi^*$ would have been the plan that is optimal with respect to $M^R$, the AI agent can choose to go with a costlier plan $\widetilde{\pi}$ (where $\widetilde{\pi}$ is still not explicable with respect to ${M}^R_h$), and provide an explanation $\mathcal{E}'$ such that $\widetilde{\pi}$ is explicable with respect to $M^R_h + \mathcal{E}'$. In \cite{Chakraborti:2018:EVE}, we show how we can synthesize behaviors that have this trade-off. 

\begin{figure}
\centering
\includegraphics[width=4in]{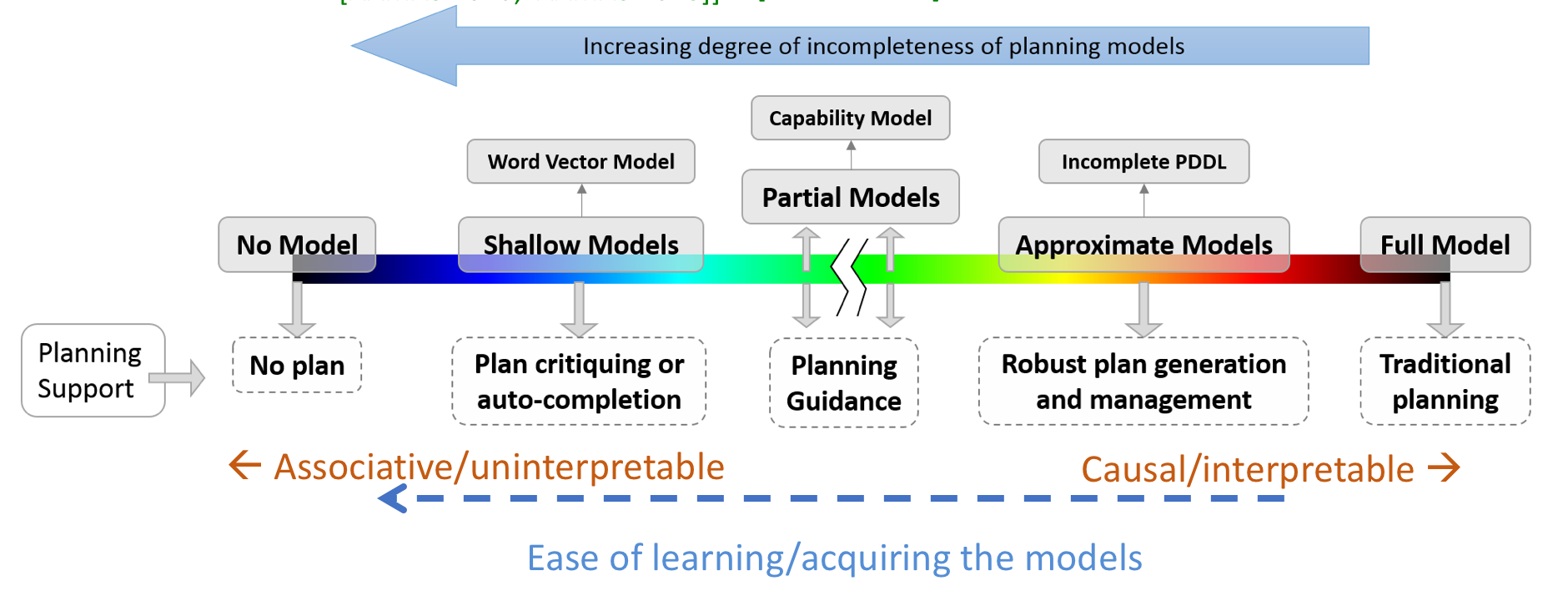}
%{usr-explaining.jpg}
\caption{\em AI agents can focus on learning a spectrum of human models--starting from fully causal specifications (e.g. PDDL) on one end to correlational/shallow models on the other}
\label{fig:spectrum}
\end{figure}

\uund{Model Acquisition}
While we focused on the question of reasoning with multiple models to synthesize explainable behavior, a closely related question is that of acquiring the models. In some cases, such as search and rescue scenarios, the human and AI agent may well start with the same shared model of the task. Here the AI agent can assume that as the default mental model. In other cases, the AI agent may have an incomplete model of the human; in \cite{sreedharan2018handling}, we provide an approach to handle the incomplete model, viewing it as a union of complete models. More generally, the AI agent may have to learn the model from the past traces of interaction with the human. Here too, the agent might get by with a spectrum of potential models--starting from fully causal specifications (e.g. PDDL) on one end to correlational/shallow models on the other (see Figure~\ref{fig:spectrum})
In \cite{tian2016discovering,Zha:2018:RPL}, we discuss some efficient approaches for learning shallow models. 
%incomplete; learn from traces
\begin{figure}
\centering
\includegraphics[width=5.5in]{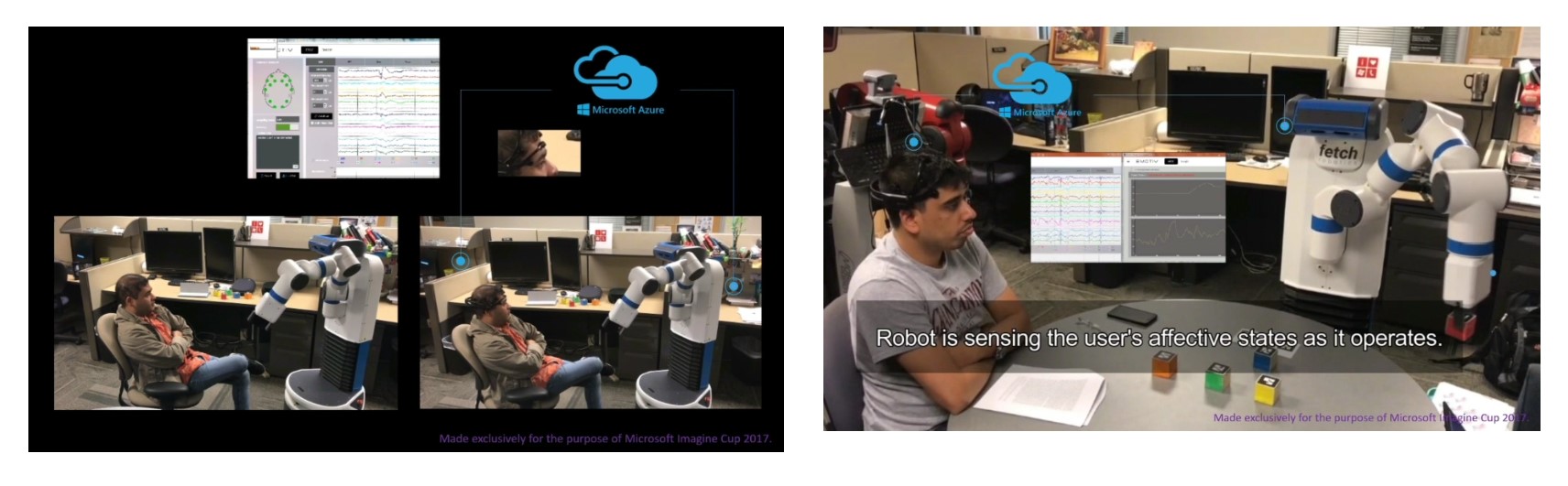}
%{usr-explaining.jpg}
\caption{\em Assessment of human affective states can be facilitated with brain-computer interface technologies (such as the Emotive helmet used here) that can supplement the normal natural communication modalities}
\label{fig:bci}
\end{figure}

\begin{figure}
\centering
\includegraphics[width=5in]{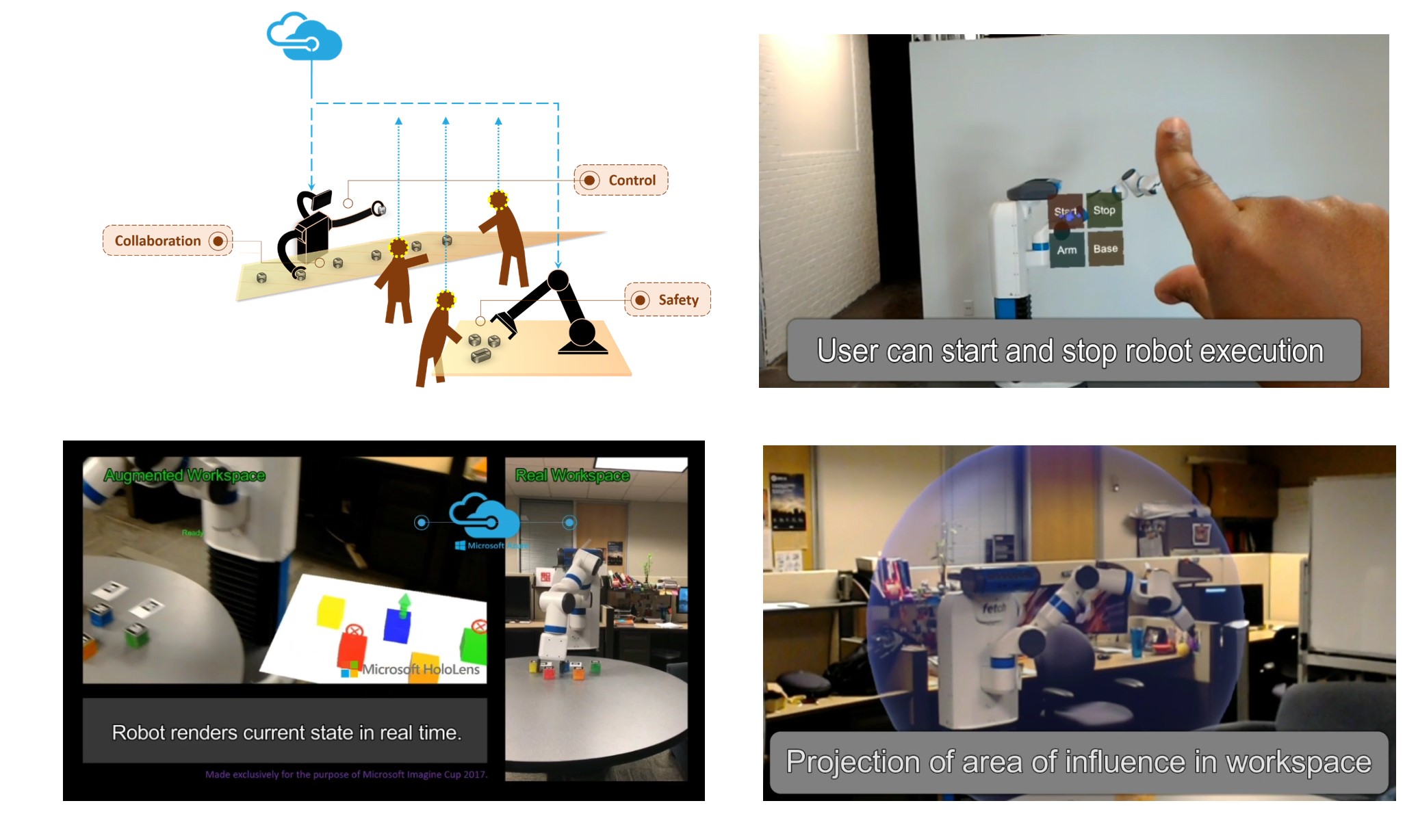}
%{usr-explaining.jpg}
\caption{\em The AI agent can project its own intentions to the human with the help of augmented reality technologies such as Hololens}
\label{fig:holo}
\end{figure}
\uund{Communicating with Humans}
Much of our work focuses on the mechanics of synthesizing explainable behavior assuming the availability of the human mental models.  A closely related problem is sensing the affective states of  human in the loop, and communicating the AI agent's own intentions to the human. This communication can be done in multiple natural modalities including speech and language and gesture recognition \cite{rehj-hri}. The human-AI communication can also be supported with the recent technologies such as augmented reality and brain-computer interfaces. Some of our own work looked at the challenges and opportunities provided by these technologies for effective collaboration. Figure~\ref{fig:bci} shows how off-the-shelf brain computer interfaces supplement natural communication modalities in assessing human affective states. Figure~\ref{fig:holo} illustrates how the agent can project its intentions with the help of augmented reality technologies such as hololens (that project the agent's intentions into human visual field). In \cite{explanatory_acts}, we look at the challenges involved in deciding when and what intentions to project. 
%workshop..

\uund{Multiple Humans \& Abstraction:}
The basic framework above can be generalized in multiple ways. In \cite{sreedharan2018hierarchical}, we show how we can handle situations where the human and AI agent have models at different levels of abstraction. In \cite{sreedharan2018hierarchical} 
%as well as \cite{sarath-foil-arxiv}, 
we consider explanations in the context of specific ``foils'' (e.g. ``{\em why not this other type of behavior?''}) presented by the humans.  In \cite{sreedharan2018handling}, we consider how the AI agent can handle multiple humans -- obviously with different models ($M^R_{h_i}$) -- in the loop, and develop the notions of ``conformant'' {\em vs.} ``conditional explanations.''

\fromrao2{after 1st: We should point out that foils allow for reduction of cognitive load for humans}

\uund{Self-Explaining Behaviors}
While the foregoing considered explanations on demand, it is also possible to directly synthesize \textit{self-explaining} behaviors. In \cite{chakraborti2018projection}, we show how the agent can make its already synthesized behavior more explicable by inserting appropriate ``projection'' actions to communicate its intentions, and also discuss a framework for synthesizing plans that takes ease of intention projection into account during planning time. 
In  \cite{explanatory_acts}, we show how we can synthesize ``self-explaining plans,'' where the plans contain  epistemic actions, which aim to shift $M^R_h$, followed by domain actions that form an explicable behavior in the shifted model. 
%while general epistemic planning is hard..

\begin{figure}[t]
\centering
\includegraphics[width=3in]{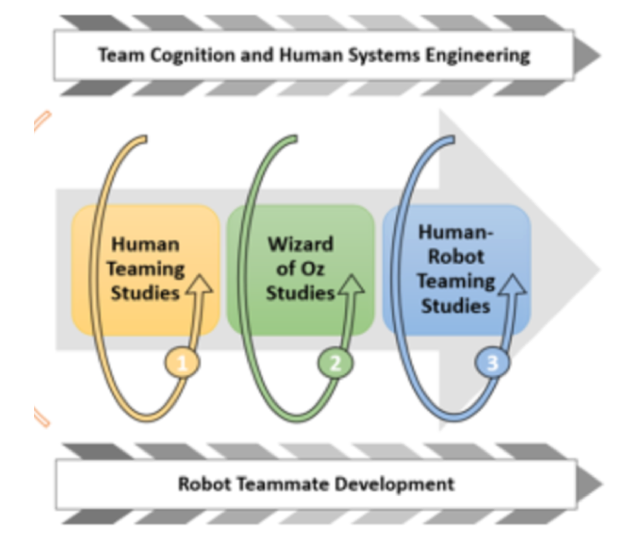}
%{usr-explaining.jpg}
\caption{\em Evaluation spirals for human-aware AI systems}
\label{fig:eval}
\end{figure}

\begin{figure}[t]
\centering
\includegraphics[width=6in]{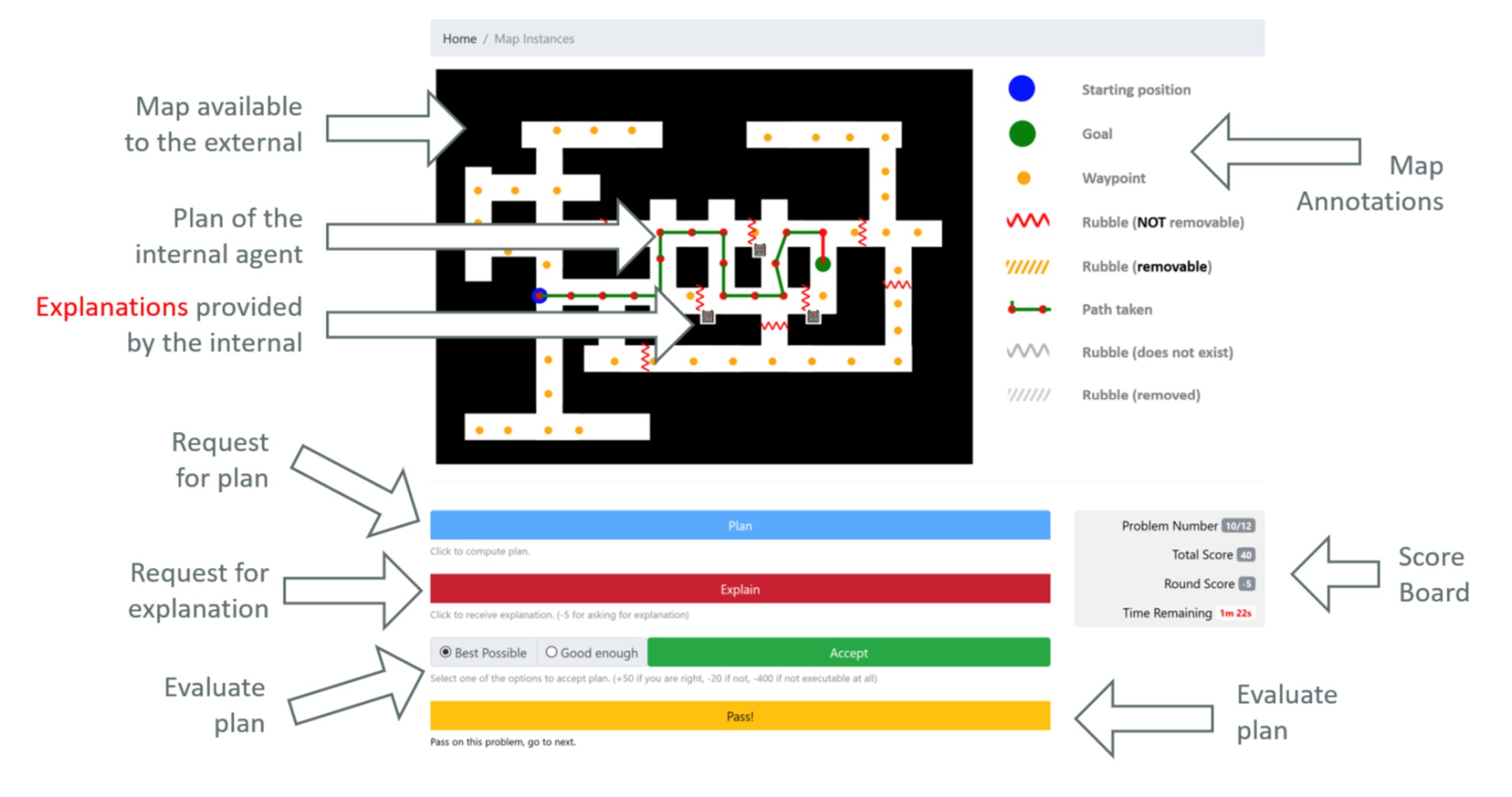}
%{usr-explaining.jpg}
\caption{\em A setup for evaluating the effectiveness of explanations produced by the AI agent in a simulated search and rescue scenario. Here the participants assumed the role of external commander and evaluated the plans provided by the AI agent. They could request for plans as well as explanations for those plans, and rate those plans as optimal or suboptimal based on that explanations (from \cite{chakraborti2019hri})}
\label{fig:hri}
\end{figure}
%evaluation and validity

%\uund{Validity \& Evaluation}
\uund{Human Subject Evaluations}
%\fromrao{Call the section Inter-disciplinary Connections \& Evaluation}
%\fromrao{need for careful evaluation}
%\fromrao{interdiciplinary human-human teaming}
An important disciplinary challenge posed by research in human-aware AI systems is that of systematic evaluation with human subjects. The temptation of a bunch of engineers unilaterally what sort of support humans will prefer should be resisted. In our own work, we collaborate with researchers in human-factors, and draw on their work on human-human teaming, as well as wizard-of-oz studies \cite{nancy-team-cognition,nancy-synthetic-teammate}. We also evaluate the effectiveness of your systems with systematic human subject studies. Figure~\ref{fig:eval} shows the evaluation spirals.  In \cite{chakraborti2019hri}, we show that people indeed exchange the type of explanations we compute, and that the need for explanations diminishes when the behavior is explicable. 

%Teaming 
%A bunch of engineers decid

%we can take some lessons from HCI

%\fromrao{cite the 20 lessons HCI paper by Dan?}

\fromrao{better title; also should it go later?}
%\und{Relations to Explainable AI (XAI) Community}
\und{Explanations, Provenance and Explainable AI}
Explainable AI (aka XAI) has become quite an active research topic recently. However, much of the work there is concerned with providing ``debugging tools" for inscrutable representations (such as those learned by deep networks for perceptual tasks), rather than as a means to human-AI collaboration. A significant part of the work in XAI is concerned with ``pointing explanations"---such as pointing the regions of an image that lead to  it being classified as an Alaskan Husky or a rare lung disease. Pointing explanations are however primitive. Imagine trying to explain/justify a decision that was made by an AI system as part of a sequential decision making scenario. Primitive  pointing explanations will  have to point to regions of {\em space-time tubes}. 
Another thread of research related to ``explanations" is providing provenance of decision. Such provenance (or certificate of correctness) is often in terms of the AI agent's own internal model and is not intended to make sense to the human in the loop. 
Model reconciliation view, in contrast, can provide explanations in terms of the features of the human and robot models of the task. They thus hew closer to psychological theories of explanation (e.g. \cite{lombrozo2006structure}). 

\fromrao{Should we put the alaskan husky and school bus examples--or would that be a bit much?}

%The explanations computed in our model reconciliation framework satisfy several desiderata---such as selectivity and contrastiveness that are seen as essential according to psychological theories.
%\cite{lombrozo2006structure}. 
%We have applied this framework in the context of human-robot interaction (e.g. \cite{chakraborti2018projection}) and interaction between humans and virtual decision support systems (e.g. \cite{sengupta2017radar}).
%We have also conducted principled human-subject studies. I

%\und{From Explanation to Manipulation}
%\und{Manipulation \& Ethical Considerations}
\und{Ethical Quandaries of Human-Aware AI Systems} 
Evolutionarily, mental modeling allowed us to both cooperate and compete with each other. After all, lying and deception are possible to a large extent because we can model others' mental states! Thus human-aware AI systems with mental modeling capabilities bring a fresh new set of {\em ethical quandaries}. 
We should also be cognizant of the fact that human's anthropomorphizing tendencies are most pronounced for emotional/social agents. After all, no one who saw Shakey for the first time thought it could shoot hoops; yet the first people interacting with Eliza\footnote{https://en.wikipedia.org/wiki/ELIZA} assumed it is a real doctor and would pour their hearts out to it (prompting Weizenbaum to abort the project!). 

Although our primary focus has been on explainable behavior for human-AI collaboration, an understanding of this also helps us solve the opposite problem of generating behavior that is deliberately hard to interpret, something that could be of use in adversarial scenarios. In \cite{kulkarni2019unified}, we present a framework for controlled observability planning, and show how it can be used to synthesize both explicable and obfuscatory behavior. 

Finally, use of mental models not only helps collaboration but also can open the door for manipulation. In principle, the framework of explanation as model reconciliation allows for the AI agent to tell white lies by bringing $M^R_h$ closer to a model different from $M^R$. For example, your personal assistant that has a good mental model of you can tell you {\em white lies} to make you eat healthy.  In \cite{chakraborti2019lie,chakraborti-xaip19}, we explore the question of whether and when it is reasonable for AI agents to lie.

\und{Epilogue}

In summary, human-aware AI systems bring in a slew of additional research challenges (as well as a fresh new set of ethical ones).  It may seem rather masochistic on our part to focus on these research challenges. As a character from Kurt Vonnegut's Player Piano remarks:
\begin{quote}
    ``If only it weren't for the people, the goddamned people," said Finnerty, ``always getting tangled up in the machinery. If it weren't for them, earth would be an engineer's paradise."
\end{quote}
On reflection however, it is easy to see that these are challenges very much worth suiting up for. After all, some of our best friends are human!

%textbf{societal impacts--lying--using mental models not only helps collaboration, but it also opens the door for manipulation.}

%\textbf{discuss need for learning--or not--aamas paper}

%\textbf{mention something about normal "why this action" queries?}

%\textbf{implementations and guis}

%\textbf{IROS projecting intentions part should go where?}

%How to connect controlled observability? 

%Should we be using widetilde M^R_h? 
%we did robots and ai-agents [radar figure?]
%extending to hierarchical models; multiple agents
%Projecting intentions.. 
\uund{Acknowledgments} 
My views on human-aware AI as well as the specific research described here was carried out in close collaboration with my students and colleagues. Special thanks to my  students Tathagata Chakraborti, Sarath Sreedharan, Anagha Kulkarni, Sailik Sengupta,  former student Karthik Talamadupula, former post-doc Yu Zhang, and colleagues Nancy Cooke, Matthias Scheutz, David Smith and Hankz Hankui Zhuo. My AAAI address as well as this write-up have benefited from the discussions and encouragement of Dan Weld, Barbara Grosz and Manuela Veloso. 
%mention barbara grosz, manuela veloso, dan weld?
%I am also grateful to the  sustained support from
Thanks also to Behzad Kamgar-Parsi,  Jeffery Morrison, Marc Steinberg and Tom McKenna of the Office of Naval Research for sustained support of our research into human-aware AI systems. Ashok Goel patiently nudged me to complete this write-up for the AI Magazine and provided helpful editorial comments. This research is supported in part by the ONR grants N00014-16-1-2892, N00014-18-1-2442, N00014-18-1-2840, the AFOSR grant FA9550-18-1-0067 and the NASA grant NNX17AD06G. 
It has been my privilege and singular honor to 
serve as the president of AAAI at a time of increased public and scientific interest in our field.  I sincerely thank the AAAI members for their trust and support. 
\fromrao{update the grants}

\bibliographystyle{plain}  % do not change this line!
%balance  % do not change this line -- unless you manually balance your last page
\bibliography{bib}  % put name of your .bib file here

%\subsection*{Bio}
%\vspace*{-0.05in}
\medskip
\noindent{\bf \large Bio:}
Subbarao Kambhampati (Rao) is a professor of Computer Science at Arizona State University. He received his B.Tech. in Electrical Engineering (Electronics) from Indian Institute of Technology, Madras (1983), and M.S.(1985) and Ph.D.(1989) in Computer Science (1985,1989) from University of Maryland, College Park. Kambhampati studies fundamental
problems in planning and decision making,
\begin{wrapfigure}{r}{2in}
\begin{center}
\includegraphics[width=2in]{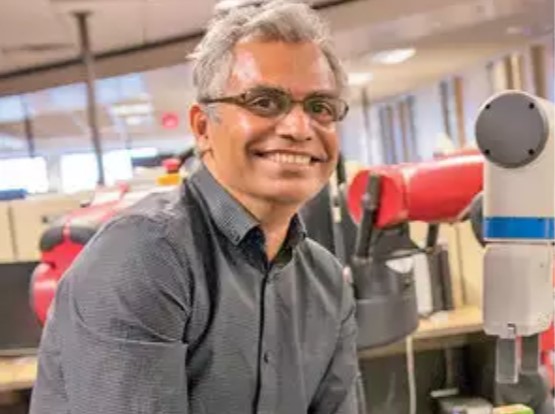}
\end{center}
\end{wrapfigure}
 motivated in particular by the challenges of human-aware AI systems.   Kambhampati is a fellow of AAAI and AAAS, and was an NSF Young Investigator. He received multiple teaching awards, including a university last lecture recognition. Kambhampati is the past president of AAAI; he served as the president of AAAI during 2016-18,  and as a trustee of IJCAI during 2013-18.  He was the program chair for IJCAI 2016, ICAPS 2013, AAAI 2005 and AIPS 2000.  He served on the founding board of directors of Partnership on AI. Kambhampati's research as well as his views on the progress and societal impacts of AI have been featured in multiple national and international media outlets. 
%URL rakaposhi.eas.asu.edu Twitter @rao2z
%\clearpage

%\textbf{implementations and guis--at least one radar paper.. connect to robot  or virtual.. }

%\textbf{IROS projecting intentions part should go where?}
\end{document}